\documentclass{article}
\usepackage{graphicx}
\usepackage{float}

\usepackage{multirow}
\usepackage{cite}
\usepackage[compact]{titlesec}
\vspace{-0.5em}
\titleformat{\section}{\large\normalsize\bfseries\filcenter}{\thesection}{1em}{\MakeUppercase}
\titleformat{\subsection}{\normalsize\bfseries}{\thesubsection.}{1em}{}
\titleformat{\subsubsection}{\normalsize\bfseries}{\thesubsubsection.}{1em}{}

\titlespacing*{\section}{-0.8pt}{*2}{*1}
\titlespacing*{\subsection}{0pt}{*1.5}{*1}
\titlespacing*{\subsubsection}{0pt}{*1}{*0.5}
\usepackage{booktabs}
\usepackage{stfloats}
\usepackage{spconf,amsmath}
\usepackage{placeins}
\usepackage{enumitem}
\setlist{nosep, leftmargin=14pt}

\usepackage{mwe} 

\title{A Dual-Mode ViT-Conditioned Diffusion Framework with an Adaptive Conditioning Bridge for Breast Cancer Segmentation}
\name{Prateek Singh$^{1}$, Moumita Dholey$^{1}$, and P.K. Vinod$^{2\star}$\thanks{$^\star$Corresponding author: vinod.pk@iiit.ac.in}}
\address{$^{1}$ iHub-Data, IIIT Hyderabad\\
         $^{2}$ CCNSB, IIIT Hyderabad, Hyderabad, 500032, India}

\begin{document}
%
\maketitle
\begin{abstract}

In breast ultrasound images, precise lesion segmentation is essential for early diagnosis; however, low contrast, speckle noise, and unclear boundaries make this difficult. Even though deep learning models have demonstrated potential, standard convolutional architectures frequently fall short in capturing enough global context, resulting in segmentations that are anatomically inconsistent. To overcome these drawbacks, we suggest a flexible, conditional Denoising Diffusion Model that combines an enhanced UNet-based generative decoder with a Vision Transformer (ViT) encoder for global feature extraction. We introduce three primary innovations: 1) an Adaptive Conditioning Bridge (ACB) for efficient, multi-scale fusion of semantic features; 2) a novel Topological Denoising Consistency (TDC) loss component that regularizes training by penalizing structural inconsistencies during denoising; and 3) a dual-head architecture that leverages the denoising objective as a powerful regularizer, enabling a lightweight auxiliary head to perform rapid and accurate inference on smaller datasets and a noise prediction head. Our framework establishes a new state-of-the-art on public breast ultrasound datasets, achieving Dice scores of 0.96 on BUSI, 0.90 on BrEaST and 0.97 on BUS-UCLM. Comprehensive ablation studies empirically validate that the model components are critical for achieving these results and for producing segmentations that are not only accurate but also anatomically plausible.

\end{abstract}
\textbf{\textit{Index Terms -}}
Breast ultrasound, lesion segmentation, denoising diffusion model, UNet, Vision Transformer, topological consistency.

\section{Introduction}

Breast cancer remains a significant global health challenge. According to GLOBOCAN 2022 estimates, there were approximately 2.3 million new cases worldwide, with Asia accounting for nearly 1 million and India alone seeing 192,000 new diagnoses. Early and accurate detection is therefore critical, and breast ultrasound (BUS) serves as a vital non-invasive imaging tool. However, interpreting BUS images is notoriously difficult due to inherent issues like speckle noise, low contrast, and ambiguous lesion boundaries, leading to time-consuming manual segmentation and inter-observer variability \cite{yap2018end}. While deep learning, particularly the UNet architecture \cite{ronneberger2015u}, has become standard for automated segmentation, traditional convolutional networks often struggle to capture sufficient global context, sometimes producing anatomically inconsistent results. This limitation has spurred interest in Vision Transformers (ViTs) for modeling long-range dependencies. Concurrently, Denoising Diffusion Models (DDMs) \cite{ho2020denoising} have emerged as powerful generative techniques, with models like MedSegDiff-V2 \cite{wu2024medsegdiff} and DPUSegDiff \cite{fan2025dpusegdiff} demonstrating remarkable fidelity in medical image segmentation by iteratively refining predictions from noise. Despite their potential, standard diffusion models face hurdles for clinical adoption: high computational costs during inference and a lack of built-in mechanisms to ensure the topological integrity of predicted structures. Addressing these gaps requires a framework that combines the global context understanding of ViTs with the generative strength of DDMs, while also prioritizing anatomical plausibility and practical efficiency. 

In this work, we introduce such a solution: a flexible, dual-mode framework based on a ViT-conditioned diffusion model. Our approach features two operational modes: (1) a high-fidelity generative mode using the full iterative sampling process, ideal for large, complex datasets, and (2) a highly efficient auxiliary mode suited for smaller datasets. In the auxiliary mode, the diffusion objective acts as a powerful regularizer, forcing the shared model backbone to learn robust anatomical representations by reconstructing clean masks from varied noise levels. This strong regularization allows a simple auxiliary head to achieve excellent segmentation performance with high inference speed, mitigating overfitting risks common in smaller datasets. \textbf{Our main contributions are threefold:}
(1) We introduce a versatile, dual-mode framework that can operate as a high-fidelity generative model or, for smaller datasets, use its diffusion objective as a deep feature regularizer to enable an efficient auxiliary prediction head. (2) We propose a novel Topological Denoising Consistency (TDC) loss, which explicitly regularizes the training process to ensure the generated masks are anatomically plausible by penalizing topological inconsistencies between denoising steps. (3) We design an Adaptive Conditioning Bridge (ACB) to effectively fuse the multi-scale global features from the ViT encoder into the UNet decoder, providing critical guidance throughout the generative process.
\begin{figure*}[htbp]
  \centering
  \includegraphics[width=0.85\textwidth]{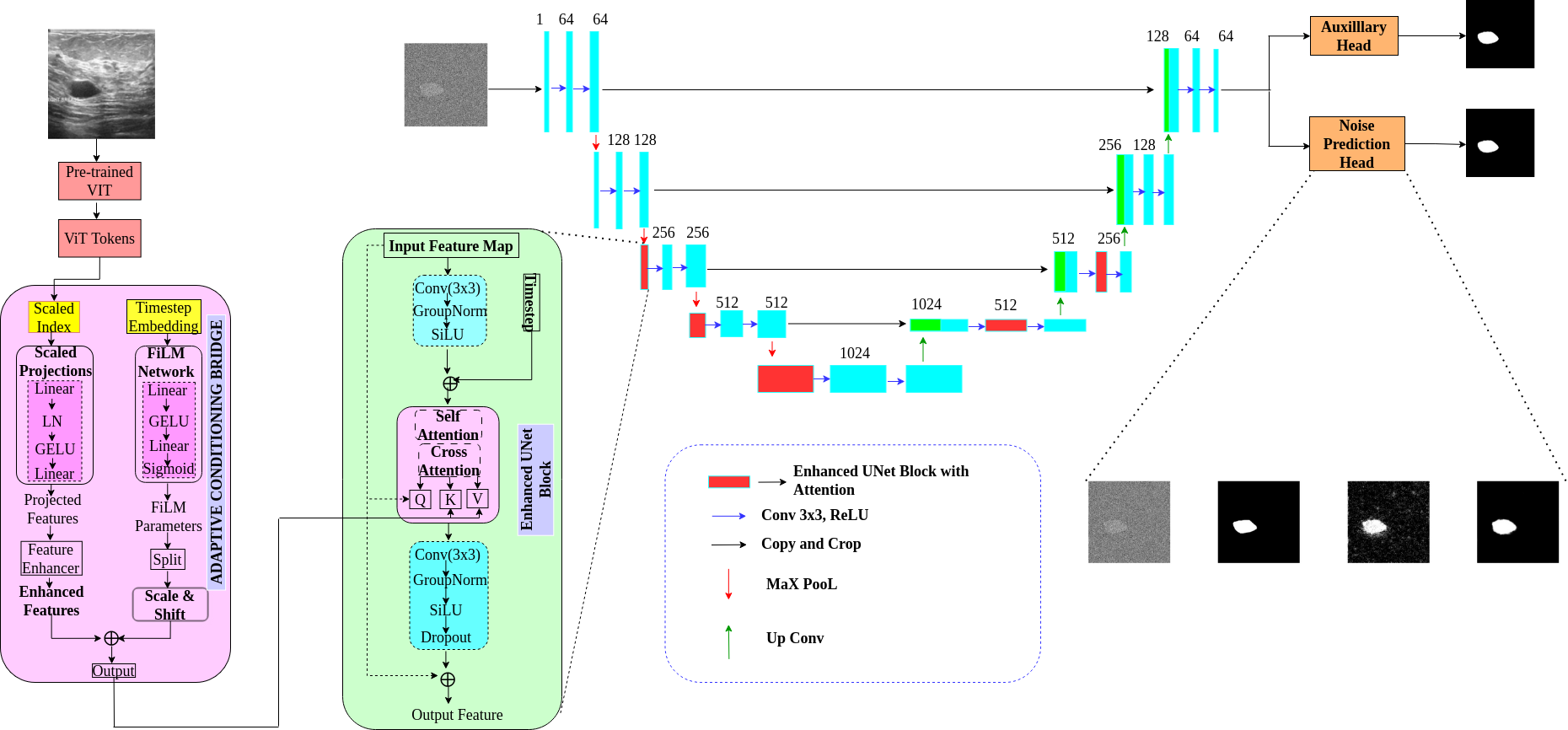}
  \caption{Architecture Overview}
  \label{fig:fullwidth_image}
\end{figure*}

\section{Methodology}

\setlength{\itemsep}{0pt}
Our framework is a flexible, conditional Denoising Diffusion Model with two modes: a rapid, regularized mode for smaller datasets and a high-fidelity generative mode for larger ones. It comprises a Conditional Diffusion UNet and a ViT-based Image Encoder.

\subsection{\textbf{Diffusion Process Formulation}}

Our method uses the DDPM framework, which has a fixed \textit{forward process} that adds Gaussian noise to a mask $x_0$ over $T$ timesteps, and a learned \textit{reverse process} that denoises it. The forward process $q$ is defined as:
{
\small
\begin{equation}
q(x_t | x_0) = \mathcal{N}(x_t; \sqrt{\bar{\alpha}_t}x_0, (1 - \bar{\alpha}_t)\mathbf{I})
\label{eq:forward_process}
\end{equation}
where $t$ is the timestep, and
{
\small
\vspace{-0.5em}
\begin{equation}
\bar{\alpha}_t = \prod_{i=1}^{t} \alpha_i
\label{eq:alpha_bar}
\end{equation}
}
is the cumulative noise product of $\alpha_i$ which represents the fraction of the original signal retained at each timestep, derived from the variance schedule $\beta_i$.
A noisy mask $x_t$ can be sampled directly:
{
\small
\vspace{-0.5em}
\begin{equation}
x_t = \sqrt{\bar{\alpha}_t}x_0 + \sqrt{1 - \bar{\alpha}_t}\epsilon
\label{eq:xt_sample}
\end{equation}
}
The \textbf{reverse process} trains a UNet $\epsilon_\theta$ to predict the noise $\epsilon$ from $x_t$, conditioned on timestep $t$ and semantic features $c$.

\subsection{\textbf{Overall Architecture}}

\setlength{\itemsep}{0pt}
As shown in Fig.1, a noisy mask $x_t$ is passed to the UNet. The UNet is conditioned by ViT features $c$ via our Adaptive Conditioning Bridge(ACB). The model has two distinct outputs:
\textbf{1.}Noise Prediction Head ($\epsilon_\theta(x_t, t, c)$) for full reverse sampling.
\textbf{2.}Auxiliary Segmentation Head ($s_{aux}$) for rapid inference.
We use a pre-trained ViT to encode the input image $I$ into a sequence of feature tokens $c=ViT(I)$, which provide conditional guidance. Our model is incredibly efficient at just 12.86M total parameters, making it 2-3.5x smaller than the MedSegDiff (25M) \cite{wu2024medsegdiff1} and MedSegDiff-V2 (46M) \cite{wu2024medsegdiff} benchmarks while outperforming them. This hybrid design is effective because we use the ViT tiny powerful global context to guide the diffusion UNet's high-fidelity generative ability. In short, the ViT identifies the "what" and "where" of the lesion, while the UNet uses that information to perfectly draw the precise mask.
\subsection{\textbf{Conditional Diffusion UNet and Adaptive Conditioning}}

Our enhanced UNet $\epsilon_\theta$ is conditioned on timestep $t$ and semantic tokens $c$.
Timestep Conditioning: The time $t$ is transformed into an embedding $e_t$ using a SinusoidalPositionEmbedding layer and a MLP.
Semantic Conditioning: The ACB converts $c$ into $c'$ using the feature enhancer and then fuses it with the enhanced UNet decoder. It generates scale ($\gamma$) and shift ($\beta$) parameters from $e_t$ and performs FiLM \cite{perez2018film}:
{
\small
\vspace{-0.5em}
\begin{equation}
\text{FiLM}(c') = \gamma \cdot c' + \beta
\end{equation}
}
These tokens are injected via cross-attention, with the UNet feature map $z$ as the query:
{
\small
\vspace{-0.5em}
\begin{equation}
\text{Att.}(z, \text{FiLM}(c')) = \text{softmax}\left(\frac{z \cdot \text{FiLM}(c')^T}{\sqrt{d_k}}\right) \cdot \text{FiLM}(c')
\end{equation}
}
\subsection{\textbf{Hybrid and Enhanced Loss Functions}}

We trained our model end-to-end with a hybrid loss function specifically designed to balance generative learning with direct segmentation. To analyze the contribution of different components, we also performed ablation studies using a separate enhanced loss.
\subsubsection{\textbf{Hybrid Loss Function}}

The \textbf{Hybrid Loss}, $\mathcal{L}_{\text{total}}$, is a weighted sum:
{
\small
\vspace{-0.5em}
\begin{equation}
\mathcal{L}_{\text{total}} = \alpha \cdot \mathcal{L}_{\text{denoising}} + \beta \cdot \mathcal{L}_{\text{Dice\_aux}} + \gamma \cdot \mathcal{L}_{\text{Focal\_aux}}
\label{eq:total_loss}
\end{equation}
}
where $\alpha, \beta, \gamma$ are hyperparameters.

The \textbf{denoising loss} $\mathcal{L}_{\text{denoising}}$ combines MSE and Dice on the implied clean mask $x'_0$:
{
\small
\vspace{-0.5em}
\begin{equation}
\mathcal{L}_{\text{denoising}} = ||\epsilon - \epsilon_{\theta}(x_t, t, c)||^2 + \lambda \cdot \mathcal{L}_{\text{Dice}}(x_0, x'_0)
\label{eq:denoising_loss}
\end{equation}
}
Simultaneously, auxiliary losses ($\mathcal{L}_{\text{Dice\_aux}}, \mathcal{L}_{\text{Focal\_aux}}$) guide the auxiliary head, tackling class imbalance and hard examples. This multi-loss setup uses the denoising challenge as strong regularization, pushing the shared UNet to learn rich, topology-aware features. The auxiliary head then leverages these features for accurate and efficient segmentation.
\subsubsection{\textbf{Enhanced Loss with Topological Denoising Consistency}}

To improve anatomical plausibility, our Enhanced Loss, $\mathcal{L}_{\text{enhanced}}$, adds a novel Topological Denoising Consistency (TDC) component with a scheduled weight $w_{\text{epoch}}$:
{
\small 

\begin{equation}
\mathcal{L}_{\text{enhanced}} = \mathcal{L}_{\text{total}} + w_{\text{epoch}} \cdot \log(1 + \mathcal{L}_{\text{topo}})
\label{eq:enhanced_loss}
\end{equation}
}
The \textbf{topological loss}, $\mathcal{L}_{\text{topo}}$, enforces structural consistency. It measures the 1-Wasserstein distance ($W_1$) between the Persistence Diagrams (PD) of the current prediction ($x'_{0|t}$) and a "look-back" prediction ($x'_{0|t-k}$):
{
\vspace{-0.5em}
\begin{equation}
\mathcal{L}_{\text{topo}} = W_1 \left( \text{PD}\left(x'_{0|t}\right), \text{PD}\left(x'_{0|t-k}\right) \right)
\label{eq:topo_loss}
\end{equation}
}
While other topological losses \cite{gupta2024topodiffusionnet} enforce adherence to a static condition, our $\mathcal{L}_{topo}$ acts as a novel \textit{stability} regularizer, ensuring anatomical plausibility by penalizing structural inconsistencies, like flickering components or holes, between successive denoising steps ($x_{0|t}^{\prime}$ and $x_{0|t-k}^{\prime}$).

\section{Experiments}

We carried out a thorough series of tests to confirm the efficacy of our suggested framework. In order to assess the contribution of each architectural and loss function innovation, we conducted thorough ablation studies and compared our model's performance to state-of-the-art (SOTA) techniques.

\subsection{Datasets and Implementation Details}

Our primary evaluation focused on the challenging domain of breast ultrasound (BUS) imaging, characterized by speckle noise and low contrast. We used three public datasets: BUSI \cite{al2020dataset} (780 images from 600 patients using LOGIQ E9 systems, categorized as 487 benign, 210 malignant, and 133 normal), BrEaST \cite{pawlowska2024curated} (256 images of benign and malignant tumors), and BUS-UCLM \cite{vallez2025bus}, comprising 683 images from 38 patients acquired with a Siemens ACUSON S2000™ system, categorized as 419 normal, 174 benign, and 90 malignant.
\begin{table}[htbp]
  \centering
  \footnotesize
  \caption{Comparison across BUSI, BReaST, and BUS-UCLM Datasets}
  
  \label{tab:comparison}
  \begin{tabular}{llcc}
    \toprule
    \textbf{Dataset} & \textbf{Methods} & \textbf{Dice} & \textbf{IoU} \\
    \midrule
    \multirow{15}{*}{\textbf{BUSI}} 
      & UNet \cite{ronneberger2015u} & 0.84 & 0.72 \\
      & TransUNet \cite{chen2021transunet} & 0.83 & 0.77 \\
      & TBConvL-Net \cite{iqbal2025tbconvl} & 0.95 & 0.91 \\
      & UNet++ \cite{zhou2018unet++} & 0.86 & 0.76 \\
      & FCN \cite{long2015fully} & 0.81 & 0.73 \\
      & DeepLabV3+ \cite{chen2018encoder} & 0.83 & 0.76 \\
      & ACSNet \cite{he2024multi} & 0.89 & 0.83 \\
      & UNet (Att.) \cite{sulaiman2024attention} & 0.85 & 0.73 \\
      & U2-MNet \cite{li2023u2} & 0.82 & --- \\
      & SaTransformer \cite{zhang2023satransformer} & 0.86 & 0.83 \\
      & Chowdary et al. \cite{chowdary2022multi} & 0.84 & 0.84 \\
      & Ours (w/ Aux. Head) & \textbf{0.96} & \textbf{0.94} \\
      & MedSegDiff \cite{wu2024medsegdiff1} & 0.65 & 0.60 \\
      & MedSegDiffV2 \cite{wu2024medsegdiff} & 0.71 & 0.67 \\
      & Ours (w/ Full Samp.) & 0.80 & 0.76 \\
    \midrule
    \multirow{6}{*}{\textbf{BReaST}} 
      & UNet \cite{ronneberger2015u} & 0.85 & 0.82 \\
      & UNet++ \cite{zhou2018unet++} & 0.86 & 0.76 \\
      & FCN \cite{long2015fully} & 0.84 & 0.76 \\
      & ACSNet \cite{he2024multi} & 0.89 & 0.83 \\
      & AnatoSegNet \cite{xu2025anatosegnet} & 0.85 & 0.75 \\
      & Ours (w/ Aux. Head) & \textbf{0.90} & \textbf{0.86} \\
    \midrule
    \multirow{6}{*}{\textbf{BUS-UCLM}} 
      & UNet \cite{vallez2025bus} & 0.69 & 0.57 \\
      & AttUNet \cite{vallez2025bus} & 0.69 & 0.58 \\
      & Sk-UNet \cite{vallez2025bus} & 0.69 & 0.58 \\
      & DeepLabv3 \cite{vallez2025bus} & 0.68 & 0.57 \\
      & Mask R-CNN \cite{vallez2025bus} & 0.77 & 0.65 \\
      & Ours (w/ Aux. Head) & \textbf{0.97} & \textbf{0.95} \\
    \bottomrule
  \end{tabular}
  \vspace{-0.5em}
\end{table}
To ensure broader applicability, we also evaluated our model on the REFUGE2 \cite{fang2022refuge2}, ISIC 2018 \cite{codella2019skin}, and BraTS 2021 \cite{baid2021rsna} datasets. The REFUGE2 dataset focuses on optic disc and cup segmentation for glaucoma assessment using retinal fundus images. ISIC 2018 provides dermoscopic images for skin lesion segmentation and classification across multiple lesion types. BraTS 2021 contains multi-modal MRI scans for brain tumor segmentation, emphasizing enhancing tumor, tumor core, and whole tumor regions. For these larger datasets, we employed the full diffusion sampling process (1000 steps) to assess the model's high-fidelity generative performance against leading transformer and diffusion-based approaches. All models were implemented in PyTorch and trained using the AdamW optimizer with a learning rate of \(5 \times 10^{-5}\), batch size of 16 and a cosine annealing schedule. 
\subsection{\textbf{Quantitative Results and Comparative Analysis}}
We compared our model against a wide range of established and recent SOTA methods. As shown in Table 1, our model establishes a new state-of-the-art on the BUSI, BrEaST, and BUS-UCLM datasets, significantly outperforming previous approaches. On the BUSI dataset, the auxiliary head achieves a Dice score of 0.96, highlighting the effectiveness of the hybrid ViT–Diffusion architecture. By contrast, the full generative sampling mode achieves a Dice of 0.80, supporting our central hypothesis: on limited datasets, the full diffusion process may overfit, whereas the auxiliary head provides a stable convergence. The model also achieves strong performance on the BReaST (Dice 0.90) and BUS-UCLM (Dice 0.97) datasets, outperforming existing transformer- and CNN-based methods. 

Extending beyond ultrasound, Table 2 summarizes performance on the REFUGE2, ISIC 2018, and BraTS 2021 benchmarks, where the model achieves consistently high Dice and IoU scores outperforming both transformer-based and diffusion-based baselines. These consistently high scores demonstrate robust cross-modality generalization, confirming the framework’s adaptability to retinal, dermoscopic, and MRI data. Overall, these results highlight the effectiveness of combining global semantic encoding (ViT) with topology-aware diffusion refinement. The framework captures both high-level anatomical structure and fine-grained boundary details, achieving stable performance across diverse imaging conditions.

\begin{table*}[htbp]
\tiny
\setlength{\tabcolsep}{3.5pt}
\renewcommand{\arraystretch}{0.9}
\centering
\caption{Comparison of State-of-the-Art Methods on REFUGE2, ISIC2018, and BraTS Datasets}
\label{tab:sota_comparison_all}

\resizebox{0.9\textwidth}{!}{%
\begin{tabular}{lcccccccc}
\toprule
& \multicolumn{4}{c}{\textbf{REFUGE2}} & \multicolumn{2}{c}{\textbf{ISIC2018}} & \multicolumn{2}{c}{\textbf{BraTS}} \\
\cmidrule(lr){2-5} \cmidrule(lr){6-7} \cmidrule(lr){8-9}
\textbf{Methods} & \textbf{Dice (Disc)} & \textbf{IoU (Disc)} & \textbf{Dice (Cup)} & \textbf{IoU (Cup)} & \textbf{Dice} & \textbf{IoU} & \textbf{Dice} & \textbf{IoU} \\
\midrule
TransUNet \cite{wu2024medsegdiff} & 95.0 & 87.7 & 85.6 & 75.9 & 89.4 & 82.2 & 86.6 & 79.0 \\
MedSegDiff-V2 \cite{wu2024medsegdiff} & 96.7 & 88.9 & 87.9 & 80.3 & 93.2 & 85.3 & 90.8 & 83.4 \\
MedSegDiff \cite{wu2024medsegdiff} & 95.1 & 87.6 & 82.1 & 72.6 & 91.3 & 84.1 & 88.9 & 81.2 \\
DPUSegDiff \cite{fan2025dpusegdiff} & --- & --- & --- & --- & 90.3 & 84.7 & 89.1 & 84.2 \\
SwinBTS \cite{wu2024medsegdiff} & 95.2 & 87.7 & 85.7 & 75.9 & 89.8 & 82.4 & 88.7 & 81.2 \\
nnUNet \cite{wu2024medsegdiff} & 94.7 & 87.3 & 84.9 & 75.1 & 90.8 & 83.6 & 88.5 & 80.6 \\
Swin-UNetr \cite{wu2024medsegdiff} & 95.3 & 87.9 & 84.3 & 74.5 & 90.2 & 83.1 & 88.4 & 81.8 \\
\textbf{Ours} & \textbf{97.2} & \textbf{92.2} & 86.9 & 75.1 & \textbf{94.1} & \textbf{86.5} & \textbf{89.7} & \textbf{84.3} \\
\bottomrule
\end{tabular}%
}
\vspace{-0.8em}
\end{table*}

\subsection{\textbf{Ablation Studies}}
We conducted ablation studies to analyze the contribution of each key component of our proposed architecture and loss functions. As shown in Table 3, removing either the diffusion-based training or the ViT conditioning led to a notable performance drop, confirming that both are indispensable for achieving optimal results.

We further assessed the contribution of our Topological Denoising Consistency (TDC) loss component. As summarized in Table~\ref{tab:combined_vertical_ablation}, integrating this loss improved both Dice/IoU scores and boundary stability, reflected in HD95(px) decreasing from 0.84 to 0.74 and reduced standard deviation. The visualizations (Fig.2) show an increase in dice score with low HD95. We also tested how a hyperparameter called "look-back distance" (k) affects the model's performance. We tested using three different k values (5,10,15). The results showed that setting k to 5 produced the highest Dice and IoU scores and the lowest error. This confirms that a smaller look-back distance improves the model's stability and overall results.
\begin{figure}[htbp]
    \centering
    \includegraphics[width=1.0\columnwidth]{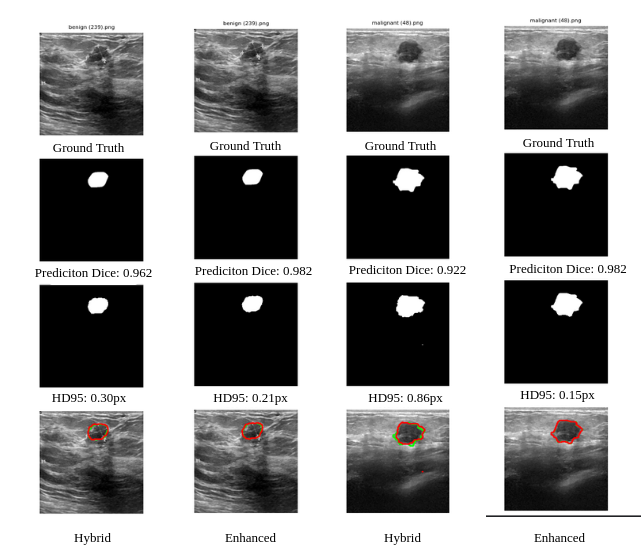}
    \caption{Hybrid vs Enhanced loss with TDC component predictions on BUSI(Red-Pred, Green-GT).}
    \label{fig:tdc_viz}
\end{figure}
\begin{table}[htbp]
  \centering
  \footnotesize 
  \caption{Ablation Study Results (Dice Score)}
  \label{tab:combined_vertical_ablation}
  \begin{tabular}{lc}
    \toprule
    \textbf{Ablation Condition} & \textbf{Dice Score} \\
    \midrule
    \multicolumn{2}{l}{\textit{Architectural Ablation}} \\ 
    Aux Head Only (No Diffusion) & 0.68 \\
    ACB OFF (Diffusion, no ViT) & 0.64 \\
    Full Model & 0.96 \\
    \midrule
    \multicolumn{2}{l}{\textit{TDC Loss Ablation (BUSI)}} \\ 
    Hybrid Loss (Baseline) & 0.96 \\
    Enhanced Loss (with TDC) & 0.97 \\
    \bottomrule
  \end{tabular}
  \vspace{-0.5em} 
\end{table}
\vspace{-0.5em} 
\section{Conclusion}
This work presents a versatile framework for medical image segmentation that combines the advantages of Vision Transformers and conditional Denoising Diffusion Models in a complementary way. Our model overcomes the narrow receptive field of traditional UNets by successfully incorporating global semantic context into the generative process through the Adaptive Conditioning Bridge (ACB), while the TDC loss enforces structural continuity, reducing artifacts such as holes and disconnections.  Ablation studies confirm that each component: the ViT encoder, diffusion backbone, ACB, and TDC loss, contributes critically to the observed performance gains. 

Unlike prior architectures such as DPUSegDiff \cite{fan2025dpusegdiff}, which rely on parallel CNN-Transformer fusion, our framework employs a more refined conditional design. Similarly, while MedSegDiff-V2 \cite{wu2024medsegdiff} introduces transformer-based conditioning through an SS-Former, it still merges features into a conventional CNN U-Net. In contrast, our model leverages a full pre-trained ViT to extract rich semantic features that guide the denoising process via FiLM-based modulation in the ACB. By directly modulating the denoising process, our framework delivers stronger and more semantically informed guidance during training. The framework’s dual-mode capability comprising a high-fidelity generative sampling mode for complex datasets and a lightweight auxiliary head for smaller ones offers flexibility between computational efficiency and precision. Despite added training costs from diffusion and topology-aware computation, the approach yields substantial gains in anatomical consistency and cross-modality generalization across ultrasound, retinal, dermoscopic, and MRI datasets. Future work will focus on accelerating sampling (e.g., consistency models) and extending to 3D and other imaging modalities for broader applicability. 
\section{Compliance with ethical standards}

\label{sec:ethics}
This is a numerical simulation study for which no ethical approval was required.

\section{Acknowledgments}

\label{sec:acknowledgments}
This work was supported by iHub-Data, International Institute of Information Technology, Hyderabad, India.

\bibliographystyle{IEEEtran}
\bibliography{references}

\end{document}